\newcommand{\CLASSINPUTtoptextmargin}{2.1cm} 
\newcommand{\CLASSINPUTbottomtextmargin}{2.95cm}
\documentclass[conference,a4paper]{IEEEtran}
\IEEEoverridecommandlockouts
\usepackage{cite}
\usepackage{verbatim}
\usepackage{amsmath,amssymb,amsfonts}
\usepackage{algorithm}
\usepackage{setspace}
\usepackage{algpseudocode}
\usepackage{graphicx}
\usepackage{textcomp}
\usepackage{xcolor}
\usepackage{tabularx}

\usepackage{adjustbox}
\usepackage{float}
\floatstyle{plaintop}
\restylefloat{table}
\usepackage{caption}
\usepackage{adjustbox}
\def\BibTeX{{\rm B\kern-.05em{\sc i\kern-.025em b}\kern-.08em T\kern-.1667em\lower.7ex\hbox{E}\kern-.125emX}}
\usepackage{longtable}
\usepackage{booktabs}
\usepackage[flushleft]{threeparttable}

\begin{document}

\title{Enhancing UAV Path Planning Efficiency Through Accelerated Learning}

\author{\IEEEauthorblockN{
Joseanne Viana \IEEEauthorrefmark{2}, 
Boris Galkin \IEEEauthorrefmark{2}, 
Lester Ho \IEEEauthorrefmark{2},
Holger Claussen \IEEEauthorrefmark{2}\IEEEauthorrefmark{3}\IEEEauthorrefmark{4}
 }

\IEEEauthorblockA{\IEEEauthorrefmark{2} Tyndall National Institute, Dublin, Ireland;}
\IEEEauthorblockA{\IEEEauthorrefmark{3} University College Cork, Ireland;} \IEEEauthorblockA{\IEEEauthorrefmark{4} Trinity College Dublin, Ireland.}
Emails: 
\{joseanne.viana, lester.ho, Boris.Galkin, holger.claussen\}@tyndall.ie\vspace{-2em} 
}
\maketitle
\begin{abstract}
Unmanned Aerial Vehicles (UAVs) are increasingly essential in various fields such as surveillance, reconnaissance, and telecommunications. This study aims to develop a learning algorithm for the path planning of UAV wireless communication relays, which can reduce storage requirements and accelerate Deep Reinforcement Learning (DRL) convergence.
Assuming the system possesses terrain maps of the area and can estimate user locations using localization algorithms or direct GPS reporting, it can input these parameters into the learning algorithms to achieve optimized path planning performance. However, higher resolution terrain maps are necessary to extract topological information such as terrain height, object distances, and signal blockages. This requirement increases memory and storage demands on UAVs while also lengthening convergence times in DRL algorithms. Similarly, defining the telecommunication coverage map in UAV wireless communication relays using these terrain maps and user position estimations demands higher memory and storage utilization for the learning path planning algorithms.
Our approach reduces path planning training time by applying a dimensionality reduction technique based on Principal Component Analysis (PCA), sample combination, Prioritized Experience Replay (PER), and the combination of Mean Squared Error (MSE) and Mean Absolute Error (MAE) loss calculations in the coverage map estimates, thereby enhancing a Twin Delayed Deep Deterministic Policy Gradient (TD3) algorithm. The proposed solution reduces the convergence episodes needed for basic training by approximately four times compared to the traditional TD3.
\end{abstract}

\begin{IEEEkeywords}
UAVs, Reinforcement Learning, TD3, Communications, Relays, Dimensionality Reduction, Memory, Storage,  5G, 6G. 
\end{IEEEkeywords}

\section{Introduction}

Unmanned Aerial Vehicle (UAV) path planning is a challenging task that involves optimizing general UAV commands and controllers according to external factors (obstacles,  telecommunication metrics, energy efficiency, security)\cite{Geraci2021}, \cite{Viana2024}, \cite{Ho24}, to achieve set goals in complex environments. 

Deep Reinforcement Learning (DRL) is increasingly being used to solve this problem, leading to significant achievements in various applications such as delivery, security, and monitoring \cite{Babatunji2023}, \cite{Fonseca2023}.  Introducing telecommunication parameters into these problems is essential for addressing coverage and network holes, but it also adds another layer of complexity to path planning. For example, calculating the wireless coverage area accurately is still a challenging task that requires precise measurements as well as extensive amounts of data (i.e., region maps, user locations, Base station locations and others), and may vary across countries due to differences in city infrastructure and telecommunication parameters \cite{3GPP23255}. Additionally, multiple UAVs can be used as relays as in \cite{Galkin2023} where inter-UAV coordination is an additional complexity. Several authors have proposed solutions which use DRL algorithms, and which include telecommunication parameters as part of path planning. Notably the evolution of such path planning algorithms is connected to the evolution of the DRL algorithms. In \cite{Susarla2020} path planning algorithms using Deep Q Learning were proposed which integrate beam tracking and path planning to achieve reasonable connectivity. In \cite{Hong2021} and \cite{Luo2024}, policy gradient algorithms including Deep Deterministic Policy Gradient (DDPG) and Twin Delayed Deep Deterministic Policy Gradient (TD3) were utilized to solve this problem. Such solutions frequently include offline training and the use of small state-spaces based on sensor inputs. These approaches may not be suitable for addressing large state-space problems, such as those related to convergence timing. Even though the authors have proposed ways to overcome long convergence time in these papers, it is well known that the convergence time increases with the size of the action and state-spaces, as the agents explore the state-space before achieving convergence.

The TD3 algorithm is a significant advancement in the field of DRL. By extending the DDPG algorithm, TD3 introduces several key components aimed at enhancing training stability and improving performance. Its twin Q-value networks, target networks, and exploration strategy contribute to more efficient learning in complex environments. Through the integration of these components, TD3 addresses common challenges such as sample inefficiency and overestimation bias, leading to more robust and reliable training outcomes. TD3 has recently been applied to UAV path planning problems as in  \cite{Hong2021}, \cite{Luo2024}. Despite its advantages, TD3 is not without limitations. One of the primary challenges associated with TD3 is its sensitivity to environmental dynamics. In tasks with non-stationary or highly stochastic environments, TD3 may struggle to adapt effectively, resulting in sub-optimal performance. Additionally, the computational complexity of TD3, stemming from its twin network architecture and multiple critics, can pose challenges in terms of resource requirements and training time. Finally, TD3 faces instability when handling large state-spaces.

In UAV wireless communications, the authors in \cite{Liu2019} propose new algorithms to overcome such limitations by using optimal subspace and graph symmetry to decrease complexity. The authors in \cite{Bozkus2022} suggest aggregating state-action pairs and combining different Markov chains formed in the wireless multiple access problem. A common insight from both papers is that large state-spaces require extensive storage, memory, and training time to produce an efficient model. A well-known technique for reducing state-spaces in Machine Learning (ML) is Principal Component Analysis (PCA), which can enhance ML performance by reducing computational costs and focusing on the most informative aspects of the data \cite{Greenacre2022}.

The objective of this paper is to optimize the size of the state-space to expedite convergence in UAV relay path planning algorithms. Specifically, we aim to improve convergence speed in path planning learning problems by applying four combined techniques: 1) Reducing the state-space size using PCA; 2) Adding a number of previous versions of new reduced state-spaces in the samples; 3) Modifying conventional TD3 by adding prioritized experience replay (PER); 4) Using the combination of Mean Squared Error (MSE) and the Mean Absolute Error (MAE) in the critic losses calculations. Through these steps, we demonstrate that faster convergence can be achieved, leading to more efficient and effective learning in UAV relay scenarios. To the best of our knowledge, we are the first to apply these techniques to a UAV relay path planning problem.

\section{System Model}

\begin{figure}[ht!]
\centering
\begingroup
 \includegraphics[scale=0.8]{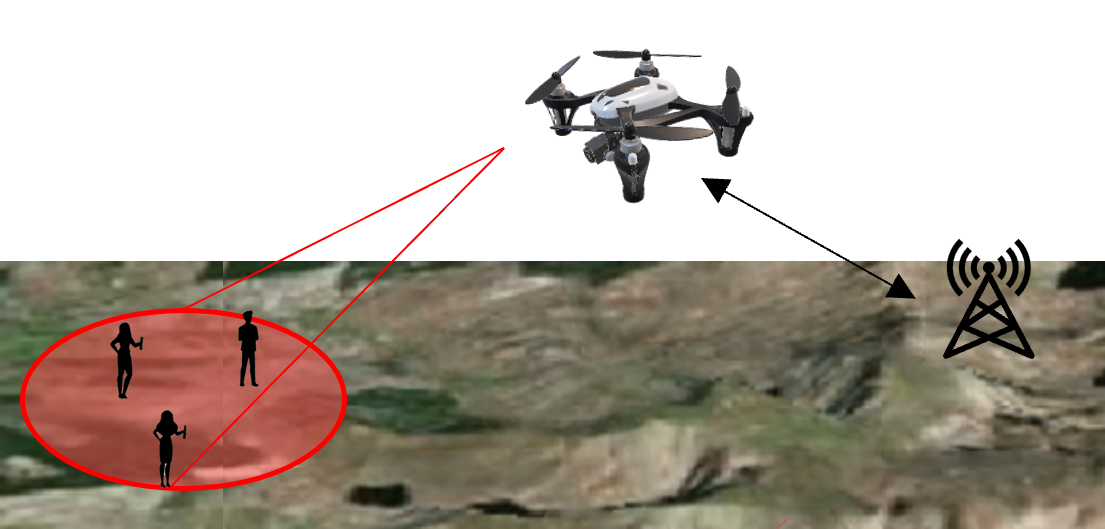 }
 \caption{Details of the air-to-ground connections in the simulation environment.}
 \label{fig:1}
\endgroup
\end{figure}

We consider a UAV assisted wireless communication system where UAVs act as radio relays to assist the communication of terrestrial users in a rural environment, who are unable to achieve the minimum quality of service from a ground base station (BS) directly. 
The relay UAV is represented as $U$
and the set of users is represented as $\mathbb{J}=\{1, 2, \ldots, J\}$.
Each user gets associated with a UAV and the UAVs then connect to the ground BS via the wireless backhaul link as presented in Fig \ref{fig:1}.
The BS-UAV and the UAV-user links are operating on orthogonal frequencies, so co-channel interference between UAV relays does not occur.

\begin{figure}[ht!]
\centering
\begingroup
 \includegraphics[scale=0.80]{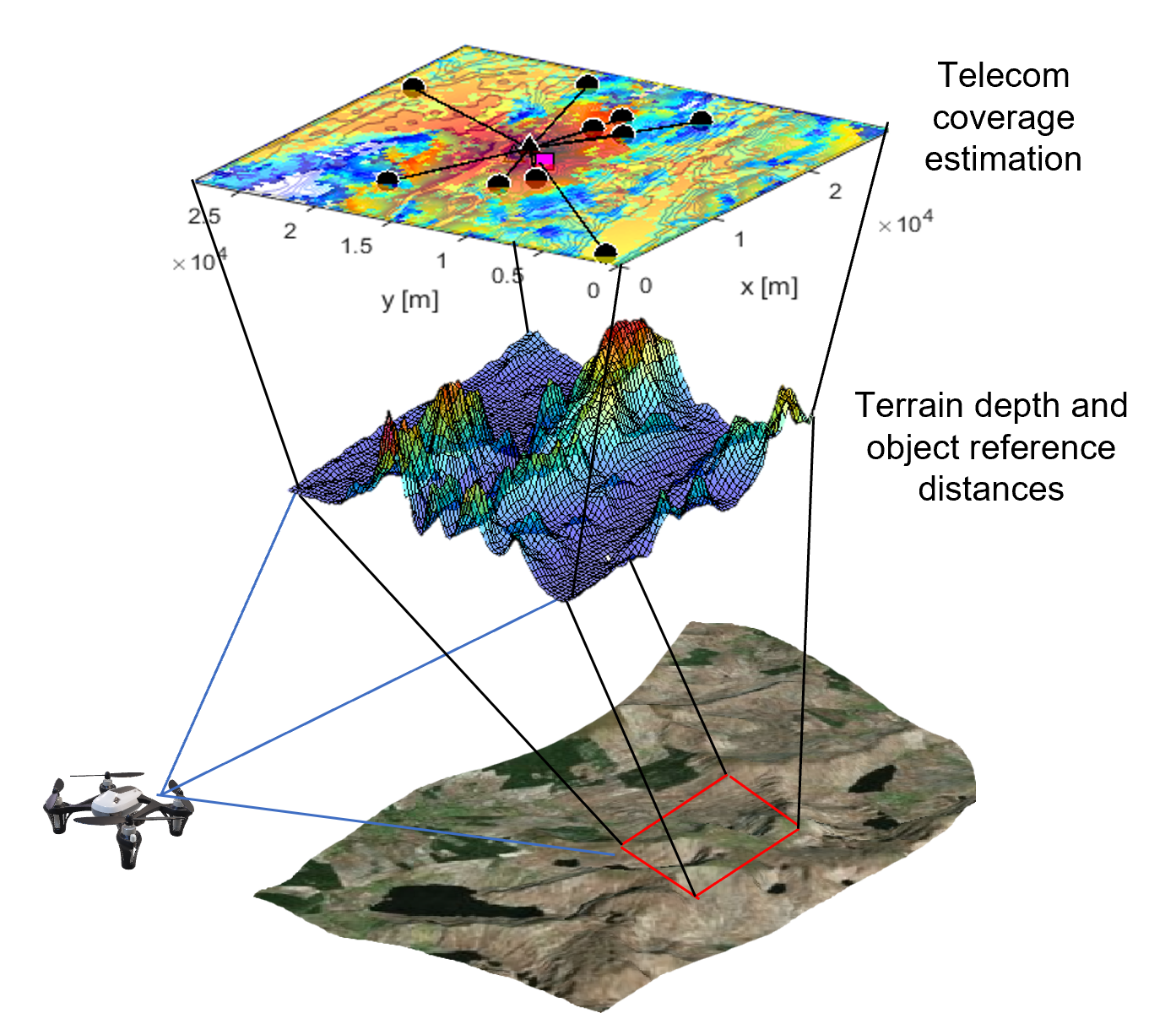}
 \caption{Details on the air-to-ground wireless propagation estimates.}
 \label{fig:2}
\endgroup
\end{figure}

We assume it is possible for the UAV relays to obtain terrain maps such as in \cite{NASA13}. 
The system knows the BS positions based on wireless connection establishment, the users locations are determined using localisation techniques, and overall relay UAV wireless coverage is determined based on measured pathloss and shadowing effects, which are used to calculate the channel propagation models.
Fig. \ref{fig:2} provides an illustration of our system and how the UAV operates. The UAV collects the terrain topology and the ground devices (i.e., BS and users) reference distances to the UAV,and estimates the air-to-ground wireless coverage.
The dots represent the users. The pink square indicates the base station location, and the triangle represents the relay UAV providing coverage to the users. The line connecting the UAV and the users signifies the wireless connection. In this scenario, all users are connected to the base station via the UAV relay.

\vspace{-3mm}
\vspace{-2mm}
\vspace{-2mm}

\subsection{Channel Model}



The channel gain between UAV $i$ and user $j$ is calculated as:
\vspace{-1mm}
\begin{eqnarray}
&&h_{{i,j}_{bs}} = PL_{{i,j}_{bs}} + KED_{i,j} + F_{j}, \label{eq:1} \\
&&h_{{i,j}_{uav}} = PL_{{i,j}_{uav}} + KED_{i,j} + F_{j}, \label{eq:2} 
\end{eqnarray}

where $PL_{i,j}$ is the distance-dependent free-space path loss, $KED_{i,j}$ is the loss due to terrain blockage which we modelled as knife-edge diffraction (KED) loss \cite{Hmamouche22}, and $F_{j}$ is the loss due to vegetation absorption at the location of user $j$.

$PL_{i,j}$ is calculated using a free-space path loss calculation, with a path-loss exponent of 2.13 \cite{allred07}:

\begin{equation}
PL_{{i, j}_{bs}}=-(20.0\log_{10}(d_{i, j})+20.0\log_{10}(f_c)-147.55). \label{eq:3} \\
\end{equation}
\begin{equation}
PL_{{i, j}_{uav}}=-(21.3\log_{10}(d_{i, j})+21.3\log_{10}(f_c)-157.2).\label{eq:4} 
\end{equation}

where $d_{i,j}$ is the distance in meters between UAV $i$ and user $j$, and $f_c$ is the carrier frequency in Hz.

$KED_{i,j}$ is used to capture the transition from line-of-sight (LOS) to non-line-of-sight (NLOS) conditions when blockage due to terrain occurs. The expressions are provided in \cite{Hmamouche22}, and are omitted here due to page constraints.

$F_{j}$ is obtained using geographical forest cover datasets \cite{SHIMADA14}, where $F_{j}$ is set to 5 dB if the user $j$ is located in a forested location, and set to zero if the user is in an open area. 
Given the UAV relays have a transmit power $P_{\text{uav-relay}}$ at a specific altitude z, the received power at user $j$ which is being covered by UAV $i$ is given as $P_{\text{user}_j} = P_\text{uav-relay} + h_{{i,j}_{uav}}$ and the received power at the uav from bs is 
$P_{\text{uav-relay}_n} = P_\text{bs} + h_{{i,j}_{bs}}$

Given that UAV relays have a transmit power $P_{\text{uav-relay}}$ at a specific altitude $z$, the received power at user $j$, covered by UAV $i$, is given by $P_{\text{user}_j} = P_{\text{uav-relay}} + h_{i, j_{\text{uav}}}$, where $h_{i, j_{\text{uav}}}$ represents the received power at user $j$ from UAV $i$. Similarly, the received power at the UAV from the base station (BS) is expressed as $P_{\text{uav-relay}_n} = P_{\text{bs}} + h_{i, j_{\text{bs}}}$, where $h_{i, j_{\text{bs}}}$ denotes the received power at the UAV from the BS.

\subsection{Coverage Estimation}

To estimate the overall coverage when a UAV is at a given horizontal location $(x, y)$, we integrate the UAV transmission power $P_{\text{uav-relay}_n}$ and the channel gains $h$ from the UAV transmissions on the ground, as defined in Eq.~\eqref{eq:1}, considering losses and blockages described by Eq.~\eqref{eq:6}.

\vspace{-1mm}
\begin{equation}
C_{xy} = P_{\text{uav-relay}} + \oint_{A}  {(h_{{uav-relay}_n)}}^{2} [\text{dB}] \label{eq:6}
\end{equation}

The integral sums up the contributions of the transmitted power over the area \( A \), accounting for terrain losses, and provides an overall estimate of the coverage area made available for users by the relay UAV. 
\vspace{-1mm}
\subsection{TD3 algorithm for Path Planning}

We are using the TD3 algorithm for UAV relay path planning. This algorithm applies online learning, allowing the agent to optimize UAV path planning by interacting with the environment and receiving rewards as feedback. The overall goal is to optimize the user coverage considering their received power $P_{\text{user}_n}$ based on the transmit power of the UAV relay $P_{\text{uav-relay}_n}$, and taking into account UAV relay locations, as well as user and base station positions.

\vspace{-1mm}
\begin{subequations}
\begin{equation}
\begin{aligned}
P1: \max \left\{ P_{\text{user}_j}, P_{\text{uav-relay}_u} \right\}, \label{eq:7}
\end{aligned}
\end{equation}
\begin{equation}
\textrm{Subject to:}  \hspace{2mm}  x_{min}  \le  x_{\text{uav}_u} \le  x_{max} \hspace{2mm} \forall x \in \mathbb{R},\label{eq:7a}
\end{equation}
\begin{equation}
 \hspace{2mm}  y_{min}  \le  y_{\text{uav}_u} \le  y_{max} \hspace{2mm} \forall y \in \mathbb{R},\label{eq:7b}
\end{equation}
\begin{equation}
 \hspace{2mm}  z_{min}  \le  z_{\text{uav}_u} \le  z_{max} \hspace{2mm} \forall z \in \mathbb{R}. \label{eq:7c}
\end{equation}
\end{subequations}
\vspace{-1mm}
Taking the above in consideration, we design the algorithm as in the following. 

\subsubsection{State Representation}

The state-space includes information such as the positions of base stations \( c_{\text{BS}} = (x_{\text{BS}1}, y_{\text{BS}1}, z_{\text{BS}1}) \), UAVs \( c_{\text{UAV}} = (x_{\text{UAV}1}, y_{\text{UAV}1}, z_{\text{UAV}1}) \), and users \( c_{\text{user}} = \{(x_{\text{user}1}, y_{\text{user}1}, z_{\text{user}1}), (x_{\text{user}2}, y_{\text{user}2}, z_{\text{user}2}), \ldots\} \), as well as environmental factors like terrain information acquired and stored in the UAV and signal strength.

The terrain information is represented as a grid or matrix denoted as \( M \). Each cell contains information about a specific region's topography, including details on tree locations and sea level.

Additionally, another matrix \( C \) contains the coverage power of the environment as calculated in equation (\ref{eq:6}). The matrix \( C \) has dimensions \( x \times y \), where \( x \) represents the x Cartesian coordinates and \( y \) represents the y Cartesian coordinates. For instance, \( C_{xy} \) signifies the coverage power at the x-coordinate and y-coordinate of matrix \( C \).

The coverage map is represented as:
\begin{align}
\mathbf{C}_{xy} = \begin{bmatrix}
C_{11} & C_{12} & \cdots & C_{1y} \\
C_{21} & C_{22} & \cdots & C_{2y} \\
\vdots & \vdots & \ddots & \vdots \\
C_{x1} & C_{x2} & \cdots & C_{xy}
\end{bmatrix} \label{eq:8}
\end{align}
In this representation, each cell \( C_{xy} \) contains information about the coverage in the corresponding area of the environment.

\subsubsection{Action Space}

The action space pertains to the 3D path planning directions for all relay UAVs. Each point within this space denotes a new $\Delta$ movement direction applied to a given UAV's current position, expressed as follows:
\vspace{-1mm}
\begin{equation}
\begin{aligned}
\text{Point} &: (x_1, y_1, z)_1 \\
\text{Point} &: (x_2, y_2, z)_2 \\
&\vdots \\
\text{Point} &: (x_n, y_n, z)_n
\end{aligned}
\label{eq:9}
\end{equation}
\vspace{-2mm}
\subsubsection{Reward Function}

The reward function is designed to reflect the objective of maximizing the area covered by the transmitting UAV relay for users connected to the wireless communication system. To expedite convergence, reward metrics associated with the overall expected user power sensitivity and the distance between the UAV and users are adopted:

\begin{itemize} 
\item {Reward 1}:
\begin{equation}
\text{r}_1 = \sum_{i=1}^{N} l_i
\end{equation}
$\text{r}_1$ defines the reward based on the feasibility of locations within the permissible operational limits where the UAV can navigate after taking a given action. $r_1$ increases when the UAV remains within the map boundaries. Here, $l_i$ refers to the coverage region indicator given by:
\begin{equation}
l_i = 
\begin{cases} 
1 & \text{if } x_{\min} \leq x_i \leq x_{\max}, \\
 & \quad y_{\min} \leq y_i \leq y_{\max}, \\
 & \quad z_{\min} \leq z_i \leq z_{\max} \\
0 & \text{otherwise}
\end{cases}
\label{eq:10}
\end{equation}
where $l_i$ determines whether the new $\Delta (x_i, y_i, z_i)$ from the action space lies within predefined coverage ranges for the UAV relay.

\item {Reward 2}:
\begin{equation}
\text{r}_2 = \frac{ \left\| \vec{d}_{\text{uav}_u, \text{final}} - \vec{d}_{\text{uav}_u, \text{init}} \right\| }{ \frac{1}{N} \sum_{i=1}^{N} \left\| \vec{d}_{\text{users}, j} - \vec{d}_{\text{uav}_u, \text{final}} \right\| }
\label{eq:12}
\end{equation} 
$\text{r}_2$ involves normalizing the UAV and user distances, providing a ratio indicator of their proximity. A positive ratio indicates the UAV moved closer to the users, resulting in a positive reward, while a negative ratio indicates the UAV moved farther away.

\item {Reward 3}:
\begin{equation}
\text{r}_3 = \frac{ \min(P_{s_{\text{users}}}) }{ \left( \frac{1}{N} \sum_{i=1}^{N} |P_{\text{users}_j} | \right) }
\label{eq:13}
\end{equation}
$\text{r}_3$ evaluates the overall available power to users relative to the minimum expected received power value $\min(P_{s_{\text{users}}})$.

\item {Total Reward}:
\begin{equation}
\text{r}_t = \text{c}_1 \text{r}_1 + \text{c}_2 \text{r}_2 + \text{c}_3 \text{r}_3
\label{eq:14}
\end{equation}
The overall reward $\text{r}_t$ for the algorithm is a weighted sum of individual rewards, with $\text{c}_1$, $\text{c}_2$, and $\text{c}_3$ denoting the respective weights.
\end{itemize}
\vspace{-1mm}
\begin{figure*}[ht]
\centering
\begingroup
 \includegraphics[scale=0.75]{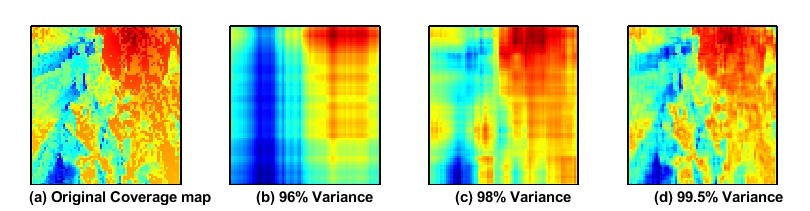}
 \caption{Coverage Map Resolution Comparison }
 \label{fig:3}
\endgroup
\end{figure*}
\vspace{-1mm}
 \section{The dimensionality reduction algorithm for State space }

The high dimensionality of the state-space associated with the coverage map \( C \), the base station, and user locations poses challenges for DRL algorithms due to the large state-space, which leads to increased computational complexity and slower convergence. To address this issue, we explore dimensionality reduction algorithms to compress the state representation while preserving relevant information. The approach is to use PCA, which identifies the principal components in the data and projects the original high-dimensional space onto a lower-dimensional subspace. By retaining the most significant features while discarding redundant information, PCA can effectively reduce the dimensionality of large state-spaces such as coverage maps. The dimensionality reduction occurs because each variance in the data provided in the maps can be associated with its covariance matrix, eigenvalues, and eigenvectors as in:
\begin{equation}
 \mathbf{M} = \frac{1}{n-1} \mathbf{C}^\top \mathbf{C} \label{eq:15}
\end{equation}
\begin{equation}
 \mathbf{E} \mathbf{v}_i = \psi_i \mathbf{v}_i \label{eq:16} 
\end{equation}
where \( \mathbf{M} \) is the covariance matrix of the coverage map, \( \mathbf{E} \) is the result of the eigenvalue decomposition, \( \psi_i \) are the eigenvalues, and \( \mathbf{v}_i \) are the eigenvectors respectively. After sorting the eigenvectors based on variance from lowest to highest, the algorithm uses them as features to estimate the coverage map.

Fig. \ref{fig:3} compares the variance percentage and the number of features needed from the PCA algorithm to achieve acceptable resolution for training. The original coverage map calculated using \eqref{eq:2} is shown in Fig. \ref{fig:3}(a), and the variance percentages with corresponding feature usage by PCA are shown in Fig. \ref{fig:3}(b)-(d): 96\% with 7\% features used in Fig. \ref{fig:3}(b), 98\% with 11\% features used in Fig. \ref{fig:3}(c), and 99.5\% with 22\% features used in Fig. \ref{fig:3}(d). We adopt the 99.5\% variance with 22\% features ratio for the results given in Section IV.

In our work, we introduce three modifications to TD3, which we refer to as Enhanced-TD3 (\text{E-TD3}). The process begins by computing the coverage map averages for each batch size and saving the overall PCA coefficients to achieve the target variance. The results from this phase correspond to an equivalent coverage map with dimension reduction. We then concatenate three additional PCA samples, referred to as "new coverage map sample," into the input and proceed with training.

In E-TD3, we apply prioritization sampling, combining Mean Squared Error (MSE) and Mean Absolute Error (MAE) in the critic loss calculations to reduce overestimation. We determine the new coordinates for the UAV relays based on the PCA components using convolutional, dense, and LeakyReLU layerslu with a negative slope of 0.01, layers as shown in Table \ref{table:1}. 

\begin{table}[ht]
\centering
\caption{Deep Neural Network Architecture}
\resizebox{\columnwidth}{!}{%
\begin{tabular}{c|p{7.3cm}}
\hline
\multicolumn{2}{c}{\textbf{Deep Neural Network Layers}} \\ \hline
\textbf{Layer Name} & \textbf{Details} \\ \hline
Conv2D Layer 1 & in\_channels=1, out\_channels=32, kernel\_size=4, stride=2, Activation: LeakyRelu \\ 
Conv2D Layer 2 & in\_channels=32, out\_channels=64, kernel\_size=2, stride=1, Activation: LeakyRelu \\
Conv2D Layer 3 & in\_channels=64, out\_channels=64, kernel\_size=1, stride=1, Activation: LeakyRelu \\
Flatten Layer 4 & Flatten \\ 
Dense Layer 5 & Dense, Units: 512, Activation: LeakyRelu \\ 
Dense Layer 6 & Dense, Units: 256, Activation: LeakyRelu \\ 
Dense Layer 7 & Dense, Units: 256 \\
Output Layer & Dense, Units: 1 \\ \hline
\end{tabular}%
}
\label{table:1}
\end{table}

\begin{algorithm}[ht!]
\caption{E\_TD3 Algorithm}
\label{alg:cap}
{\fontsize{9.5}{9.5}\selectfont
\begin{algorithmic}[1]
\Require $ C_{x-3, y-3}$, $C_{x-2, y-2}$, $C_{x-1, y-1}$, $C_{xy} $, $J_n(x, y, z)$, $b_n(x, y, z)$
\State \textbf{Step 1: Data Preparation}
\State Calculate PCA elements from the required data.
\Ensure $C_{pca-3}$, $C_{pca-2}$, $C_{pca-1}$, $C_{pca}$
\State Concatenate $C_{pca-1}$, $C_{pca-2}$, $C_{pca-3}$, $C_{pca}$ to form the input data. 
\State \textbf{Step 2: Prioritize TD3}
\State Initialize replay buffer $B$.
\State Initialize critic networks $Q_1$, $Q_2$, and actor network $\pi$ with random parameters.
\For{each iteration}
 \State Sample mini-batch of transitions $(s, a, r, s')$ from $B$.
 \State Compute target actions with noise: $a' = \pi(s') + \epsilon$.
 \State Compute target Q-values: $\hat{y} = r + \gamma \min(Q_1(s', a'), Q_2(s', a'))$.
 \State Compute critic losses:
 \[
L_{\delta}(Q_{x}, \hat{y}) =
\begin{cases} 
\frac{1}{2} (Q_{x} - \hat{y})^2 & \text{if } |Q_{x} - \hat{y}| \le \delta, \\
\delta \left(|Q_{x} - \hat{y}| - \frac{1}{2} \delta\right) & \text{if } |Q_{x} - \hat{y}| > \delta. 
\end{cases}
 \] 
\State Where $x$ represents each critic network $Q_1$, $Q_2$.

 \State Update critics $Q_1$, $Q_2$ by minimizing $L_{\delta}$.
 \If{every $d$ steps}
 \State Update actor $\pi$ by the deterministic policy gradient:
 \[
 \nabla_{\theta} J = \frac{1}{N} \sum \nabla_a Q_1(s, a) \big|_{a=\pi(s)} \nabla_{\theta} \pi(s).
 \]
 \State Update target networks:
 \[
 \theta' \leftarrow \tau \theta + (1 - \tau) \theta'.
 \]
 \EndIf
\EndFor
\State \textbf{Step 3: Update Policy}
\Ensure $A = \text{Point}_1: (x_1, y_1, z_1)$
\end{algorithmic}
}
\end{algorithm}

The complexity of data manipulation depends on how well we can integrate all data within the deep network so that consecutive layers can extract relationships between input values.
, we do that after the convolutional layers where we combine the information from the elements localization in the overall estimated data.
The new coverage map sample enters the convolutional layers. 
After the estimation in the convolutional layers, we concatenate the results with all vectors related to the positions in the system into a unique vector and process the data in the linear layers. This process happens in both critic and actor network. 
The deep network architecture in the actor and in the critic are similar across all layers. 
Algorithm \ref{alg:cap} illustrates the details of the changes proposed in E-TD3, where $C$ is the coverage map sample, and $J_n(x, y, z)$, $b_n(x, y, z)$ represents the users and base station coordinates samples. We concatenate the $C_{pca-1}$, $C_{pca-2}$, $C_{pca-3}$, $C_{pca}$ samples and execute the \text{E-TD3} with Mean Squared Error (MSE), the Mean Absolute Error (MAE) loss calculations and prioritization (PER). 

\section {Results}

This section presents the results and insights gained from our UAV experiment using \text{TD3}, \text{TD3+PCA}, and \text{E-TD3}, along with details of the data used to train all the algorithms. Unless explicitly stated , the general network and DRL parameters employed in the experiment are described in Tables \ref{table:2} and \ref{table:3}, respectively. 

\begin{table}[ht]
\centering
\begin{threeparttable}
\begin{tabular}{@{}ll@{}}
\toprule
Scenario Parameters & Values \\ \midrule
Terrestrial Users & 15 \\
Relay UAVs & 1 \\
Relay height & 10 - 300 m \\
Users height & 1.5m \\
BS height & 10m \\
BS power & 36.98 dBm \\
UAV power & 36.98 dBm \\
Users position & URD* \\
Base Station position & URD* \\
Area Coverage Maps  & 28000 vs 28000 $m^2$ \\
Max Distance between users &  $10000 m$ \\

\bottomrule
\end{tabular}
 \begin{tablenotes}
 \small
 \item *URD - Uniformly Random Distributed.
 \end{tablenotes}
 \end{threeparttable}
\caption{Network Parameters.}
\label{table:2}
\end{table}
\vspace{-1mm}
\begin{table}[ht]
\centering
\begin{tabular}{lllccc}
\toprule
& \text{TD3} & \text{TD3-PCA}  & \text{E-TD3}\\
\midrule
Actor Learning Rate & 1e-3 &  1e-3  & 1e-5 \\ 
Critics Learning Rate&  4.0e-4 & 5e-4& 6.4e-4\\
Discount Factor &0.99 & 0.95  & 0.95\\
Batch Size & 100 & 100  & 100 \\
Soft update factor &0.005&  0.005& 0.005\\
\bottomrule
\caption{Reinforcement learning parameters}
\label{table:3}
\end{tabular}
\end{table}
\vspace{-2mm}
\subsection {Convergence Rate}
\begin{figure}[ht]
\centering
 \includegraphics[scale=0.61]{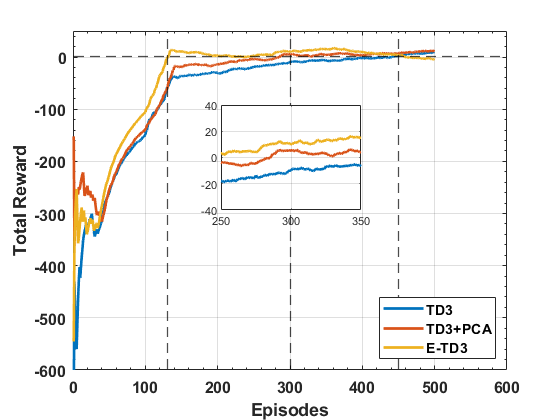}
 \caption{100 average training comparison after running 500 episodes. (obtaining a reasonable training score using TD3, TD3+PCA, E-TD3)}
 \label{fig:4}
\end{figure}

\vspace{-1mm}
Figure \ref{fig:4} compares the learning curves of \text{TD3}, \text{TD3+PCA} and \text{E-TD3} for a fixed users scenario. We randomly simulate the system 100 times, train across 500 episodes in each simulation, and record the average rewards across all steps for each episode. 

\vspace{-1mm}

As shown in Fig. \ref{fig:4}, all the comparative algorithms converge quickly. \text{E-TD3} achieves fast convergence after approximately 120 episodes,  \text{TD3+PCA} achieves similar rewards after 300 episodes and \text{TD3} converges after 450 episodes. The rapid convergence can be attributed to the dimensionality reduction, timed compression to different samples, and decayed noise used in \text{E-TD3}. First, decreasing the state-space's dimensionality using the $pca\_scores$ and combining previous timed samples of it with the average mean in the $batch\_size$ effectively accelerates the convergence rate. We chose three previous samples to stay within our overall coverage map, as dimensionality reduction provides an encoded map with a size of 22\% of the general map. 
Second, with dynamic $\epsilon$ decay in exploration, data utilization efficiency is guaranteed. At the early stage of the training process, more explorations are attempted. With more training episodes, the proportion of exploration noise decreases, and UAVs make decisions mainly based on the trained policy with exploitation.
As a result, \text{E-TD3} obtains more rewards than its comparative counterparts. In this case, the priority is given to learning the best paths, but the accuracy of learning the patterns in the coverage is also relevant.  

\begin{figure}[ht]
\centering
\begingroup
 \includegraphics[scale=0.63]{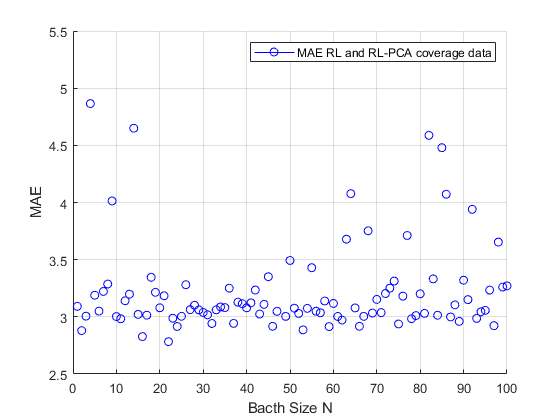}
 \caption{MAE Between the Original and Dimensionality Reduced Coverage Maps in one Batch Size }
 \label{fig:5}
\endgroup
\end{figure}

\subsection {Comparison between coverage maps in the batch size}

Fig. \ref{fig:5} presents the mean absolute error (MAE) between non-PCA and PCA coverage maps for one $batch\_size$.A lower MAE signifies a higher similarity between the coverage maps, with a value close to 0 in db indicating a close match between the overall maps and  higher values suggests complete dissimilarity. 

 With this comparison, we conclude that even though the we use only 22\% of the original coverage map available encoded with pca and the average scores per $batch\_size$, the coverage maps are similar to each other the average difference in each coverage map on the batch size is approximately 3db, such that we are able to achieve good UAV positioning while using lower dimensions.

\section{Conclusions} 

In this paper we presented a dimensionality reduction learning algorithm for target coverage estimation in UAV wireless communication relays. The algorithm uses TD3 architecture and two proposed extensions. The results showed that the best-accomplished performance was achieved by our proposed  \text{E-TD3}, where the coverage maps were reduced via PCA, and previous timed samples were compressed and used for training. This technique reveals that training with dimensionality reduction-added samples and changes in the loss estimations using harmonic averages is more effective for fast convergence and learning. Furthermore, we show that although PCA-encoded data uses less information for training, the recovered coverage maps are statistically similar to the overall coverage maps, which is fundamental in UAV relay scenarios.

\section{Future Work} 

In future research, we plan to extend our reinforcement learning implementation in UAV scenarios. Key areas of focus will include adapting the algorithm for multi-UAV coordination to enhance network coverage and reliability. We also aim to incorporate dynamic environment modeling to handle real-time changes in obstacles or user locations. Finally, we intend to validate our approach through real-world experiments. These advancements will contribute to more robust and efficient UAV-based wireless communication systems.

\section{Acknowledgements} 
This work was funded by Science Foundation Ireland under the Mistral project grant number 21/FIP/DO9949.


\vspace{9pt}
\end{document}